\pdfoutput=1

\documentclass[letterpaper, 10 pt, conference]{ieeeconf}

\IEEEoverridecommandlockouts
\usepackage{graphicx}
\usepackage{tikz}
\usepackage{amsmath}
\usepackage{amssymb}
\usepackage{mathtools}

\usepackage{amsthm}
\usepackage{booktabs}
\usepackage{multirow}
\usepackage{algorithm}
\usepackage{algpseudocode}
\usepackage{cite}
\usepackage{url}

\theoremstyle{plain}

\theoremstyle{definition}

\title{\LARGE \bf
SenseFuse: Label-Free Fusion of Image and Shape Encoders for Open-Vocabulary 3D Instance Segmentation
}

\author{Euiseok Han$^{1}$, Tri Ton$^{1}$, Hwanhee Kim$^{1}$, Seungyeon Ryu$^{1}$, and Chang D. Yoo$^{1}$%
    \thanks{Korea Advanced Institute of Science and Technology (KAIST). (email: hanes1207@kaist.ac.kr; tritth@kaist.ac.kr; khhandrea@kaist.ac.kr; sylucyryu@kaist.ac.kr; cd\_yoo@kaist.ac.kr)}
}

\makeatletter
\long\def\@makecaption#1#2{%
  \ifx\@captype\@IEEEtablestring
    \centerline{\footnotesize #1}%
    \vskip 1.5pt
    \begingroup\footnotesize
      \setbox\@tempboxa\hbox{#2}%
      \ifdim\wd\@tempboxa>\hsize
        \noindent\unhbox\voidb@x #2\par
      \else
        \centerline{\box\@tempboxa}%
      \fi
    \endgroup
    \@IEEEtablecaptionsepspace
  \else
    \@IEEEfigurecaptionsepspace
    \setbox\@tempboxa\hbox{\footnotesize #1.~~ #2}%
    \ifdim\wd\@tempboxa>\hsize
      \setbox\@tempboxa\hbox{\footnotesize #1.~~ }%
      \parbox[t]{\hsize}{\footnotesize\noindent\unhbox\@tempboxa#2}%
    \else
      \ifcenterfigcaptions \hbox to\hsize{\footnotesize\hfil\box\@tempboxa\hfil}%
      \else \hbox to\hsize{\footnotesize\box\@tempboxa\hfil}%
      \fi
    \fi
  \fi}
\makeatother

\begin{document}

\maketitle
\thispagestyle{empty}
\pagestyle{empty}

\begin{abstract}
    Open-vocabulary scene understanding is fundamental for robotics, laying the groundwork for spatial reasoning and object manipulation.
    While closed-vocabulary 3D instance segmentation heavily leverages 3D shape information, state-of-the-art open-vocabulary methods remain predominantly restricted to 2D image features or image-distilled representations during mask labeling.
    In this paper, we propose SenseFuse, a label-free fusion method that balances 2D image and 3D shape encoders for robust open-vocabulary 3D instance segmentation, refining only the mask-labeling stage of existing pipelines.
    We reveal that 2D image and 3D shape encoders exhibit largely disjoint failure patterns and rarely share identical wrong labels, whereas two 2D image encoders frequently repeat the same errors. This distinct behavior makes the 2D and 3D pair inherently complementary.
    We introduce an adaptive mechanism that selects a scene-level fusion weight to maximize a label-free sensitivity measure, estimated directly from a single scene's unlabeled proposals in milliseconds.
    SenseFuse improves labeling accuracy in every evaluated setting across ScanNet200, Replica, and ScanNet++, recovering 67--100\% (median 93\%) of the gain achievable with an oracle weight, and it raises instance AP in 21 of 22 reported settings.
    Code is available at \url{https://github.com/hanes1207/SenseFuse}.
\end{abstract}

\section{Introduction}
\label{sec:introduction}

\begin{figure}[t]
\centering
\includegraphics[width=\columnwidth]{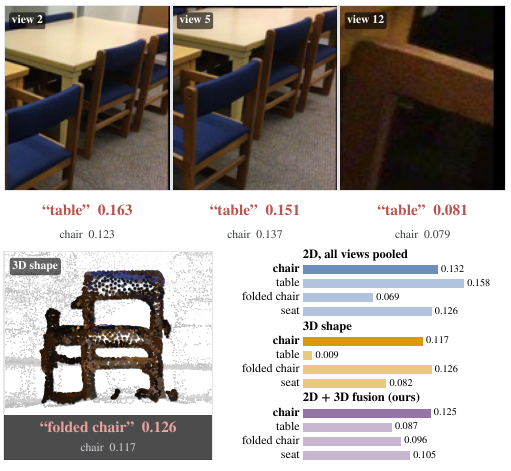}
\caption{\textbf{SenseFuse}
    improves instance labeling by linearly fusing 2D image and 3D shape scores at a label-free estimated weight.
    We show a failure case of Open-YOLO3D: multi-view crops, per-class scores of both heads (centered cosine scores, Eq.~\eqref{eq:scores}), and the fused verdict (solid bars: ground truth classes).
    The 2D and 3D heads misread the chair as ``table'' and ``folded chair'', respectively --- yet their errors disagree. 
    Fusing the two score vectors recovers ``chair''.
}
\label{fig:cover}
\vspace{-3mm}
\end{figure}

In robotics, scene understanding is a pivotal problem and a prerequisite for most downstream tasks.
Both object manipulation~\cite{liu2024okrobot} and spatial reasoning~\cite{cheng2024spatialrgpt} rely on identifying target objects in a scene, driving rapid progress in 3D Instance Segmentation (3DIS)~\cite{yilmaz2026Volt} on benchmarks such as ScanNetv2~\cite{dai2017scannet}.
Closed-vocabulary 3DIS methods inherently exploit explicit 3D shape information~\cite{mask3d,yilmaz2026Volt,jiang2020pointgroup,vu2022softgroup,ngo2023isbnet}. 
However, their fixed category sets fail on novel classes, motivating open-vocabulary 3DIS~\cite{peng2023openscene,ding2023pla} in both training-based~\cite{ding2023pla,yang2023regionplc,ding2024lowis3d} and training-free~\cite{takmaz2023openmask3d,nguyen2024open3dis} forms. 
Yet most open-vocabulary pipelines label masks from 2D image encoders or image-distilled representations \cite{radford2021clip}, leaving 3D shape evidence unused during label assignment. 
As Fig.~\ref{fig:cover} illustrates, 2D image encoders fail in predictable ways: a chair occluded by a table is misidentified as a table in most 2D views --- and the 3D head \cite{zhou2023uni3d} errs too, yet differently, which is precisely what fusion can exploit.
How to effectively leverage point clouds for 3D shape evidence at the label-assignment stage of open-vocabulary 3DIS remains an open problem.

The primary difficulty is achieving zero-shot generalization without class bias, because training 3D networks on 2D pseudo-labels heavily skews them toward seen categories~\cite{ding2023pla}. 
We instead propose SenseFuse, a label-free fusion module that balances a frozen 2D image encoder against a frozen 3D shape encoder at the scene level. 
Because it operates directly on candidate mask proposals, a robot entering an unseen environment can recalibrate its instance labels without additional supervision or retraining.

Our starting point is empirical. 
The two modalities err on largely disjoint instances and rarely share identical wrong labels, whereas two 2D image encoders repeat each other's mistakes far more often than not.
We quantify each head's reliability by its sensitivity --- how far it lifts the correct class above its own background noise --- and fuse the two heads with the single scene-level weight that maximizes the combined sensitivity, computable in closed form from the first two score moments alone.
Under Gaussian scores, this rule exactly matches the Bayes decision rule.
Integrated with Open-YOLO3D~\cite{boudjoghra2024openyolo3d}, OpenMask3D~\cite{takmaz2023openmask3d}, and Open3DIS~\cite{nguyen2024open3dis} on ScanNet200~\cite{rozenberszki2022language}, Replica~\cite{straub2019replica}, and ScanNet++~\cite{yeshwanth2023scannet++}, fusing with Uni3D~\cite{zhou2023uni3d} at our predicted weights recovers the vast majority of the oracle gain, with weight estimation averaging $3.5$\,ms per scene.

\begin{figure*}[t]
\centering
\includegraphics[width=\textwidth]{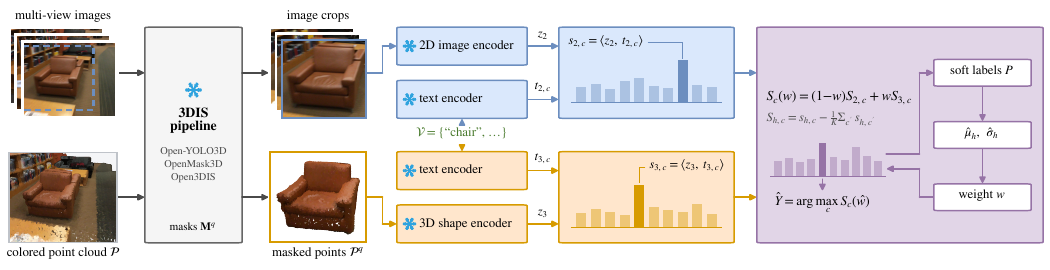}
\caption{\textbf{Overview of the SenseFuse pipeline.}
    A frozen 3DIS pipeline (Open-YOLO3D, OpenMask3D, or Open3DIS) provides mask proposals $\mathbf{M}^{q}$, image crops, and masked point clouds $\mathcal{P}^{q}$.
    The 2D image and 3D shape encoders then compute class scores as cosine similarities between these representations and each head's text anchors.    
    The two score spaces are row-centered and linearly fused with a single scalar weight $w$.
    Soft responsibilities computed from the fused scores yield per-head reliability moments $(\hat\mu_h, \hat\sigma_h)$, which update $w$ until convergence at $\hat w$ --- one scalar per scene, estimated from unlabeled masks alone --- and the final label is the $\arg\max$.
    No component is trained, and the mask geometry is untouched.
}
\label{fig:architecture}
\vspace{-3mm}
\end{figure*}

Our contributions can be summarized as follows:
\begin{itemize}
    \item We show that a 3D shape encoder and a 2D image encoder fail on largely disjoint instances and rarely share identical wrong labels, whereas two 2D image encoders repeat each other's errors. This makes the 2D--3D pair, rather than a second 2D image encoder, the profitable fusion target.
    \item We derive a closed-form, scene-level fusion weight from label-free sensitivity estimates in a centered cosine-score space. This calibrates on a single scene's unlabeled proposals in milliseconds, requiring no training and no ground-truth labels.
    \item Across three benchmarks and three baseline pipelines, SenseFuse improves labeling accuracy in every setting (median $+7.1$ points), recovering a median $93\%$ ($67$--$100\%$) of the gain available at an oracle weight.
\end{itemize}
\section{Related Work}
\label{sec:related_work}
 
\subsection{Open-Vocabulary 2D Scene Understanding}

Large-scale image--text pretraining, most notably CLIP~\cite{radford2021clip} and ALIGN~\cite{jia2021align}, enables recognition beyond a fixed taxonomy by aligning visual features with a shared vision--language embedding space.
Building on this alignment, open-vocabulary detection and semantic segmentation ground region features to text embeddings~\cite{zhong2022regionclip,zhou2022detic,liu2024grounding,cheng2024yoloworld,li2022lseg,ghiasi2022openseg,liang2023ovseg,xu2022groupvit,xu2023odise}, instance segmentation predicts per-instance masks via cross-modal pseudo-labeling~\cite{huynh2022ovis}, and class-agnostic segmenters such as SAM~\cite{kirillov2023sam} supply masks for a vision--language model to classify.
These models generalize remarkably to novel classes, but the appearance cues they rely on are precisely what degrade under the viewpoint shifts, occlusions, and scale ambiguities of real 3D environments.
Our method therefore treats 2D predictions not as the sole labeling signal but as one of two complementary evidence sources.
 
\subsection{Closed-Vocabulary 3D Scene Understanding}
Fully-supervised 3DIS methods predict masks and labels from a predefined category set, inherently exploiting 3D shape and geometry---whether by bottom-up point grouping~\cite{jiang2020pointgroup,vu2022softgroup}, proposal-based dynamic kernels~\cite{ngo2023isbnet}, or transformer architectures that refine instance queries~\cite{mask3d}, unify segmentation tasks~\cite{oneformer3d}, or revisit vanilla attention for 3D~\cite{yilmaz2026Volt}.
Additionally, ODIN~\cite{jain2024odin} explicitly fuses 2D and 3D features within a single model.
Despite strong geometric sensitivity on benchmarks such as ScanNetv2, their fixed label set prevents generalization to novel categories.
SenseFuse retains this geometric awareness through a 3D shape encoder while removing the closed-set constraint entirely.
 
\subsection{Open-Vocabulary 3D Scene Understanding}
Open-vocabulary 3D methods follow two scene-level paradigms---training-based distillation and training-free instance-level segmentation. A parallel line of research develops text-aligned 3D shape encoders.
Training-based approaches project 2D vision--language features onto points~\cite{peng2023openscene}, or they align 3D features with text captions at the point~\cite{ding2023pla,yang2023regionplc} or instance level~\cite{ding2024lowis3d}.
These methods train dedicated 3D backbones to produce proposals and labels end-to-end. However, their reported degradation on unseen categories~\cite{ding2023pla} motivates our training-free approach.
Training-free instance-level methods instead label class-agnostic proposals with frozen 2D models.
OpenMask3D~\cite{takmaz2023openmask3d} utilizes multi-scale CLIP crop embeddings, and Open3DIS~\cite{nguyen2024open3dis} merges superpoints under multi-view 2D masks to boost recall on small objects.
Alternatively, Open-YOLO3D~\cite{boudjoghra2024openyolo3d} replaces heavy feature lifting with a fast open-vocabulary 2D detector, while OVIR-3D~\cite{lu2023ovir3d} back-projects 2D instance masks onto the point cloud.
Across their diverse proposal strategies, these pipelines assign labels almost exclusively from 2D image encoders or image-distilled representations. Native 3D shape evidence is left completely unused at the label-assignment stage, causing critical failures precisely where depth or occlusion dominate.
In parallel, text-aligned 3D shape encoders have matured independently.
ULIP~\cite{xue2023ulip,xue2024ulip} aligns point cloud, image, and text encoders in a shared space.
OpenShape~\cite{liu2023openshape} and Uni3D~\cite{zhou2023uni3d} scale this recipe for strong zero-shot shape recognition, scoring point clouds directly against class names without 2D projection.
Yet, because they are trained on object-centric shapes, they underperform 2D image encoders in cluttered scenes on their own (Sec.~\ref{sec:experiments}).
SenseFuse bridges these two lines of work. It neither generates proposals nor trains a backbone. Instead, it re-evaluates the class assignments from an existing pipeline, rebalancing 2D appearance against a native 3D shape encoder through a label-free fusion calibrated per scene.

\section{Method}
\label{sec:method}

To exploit 3D shape information, we balance two scoring heads--a 2D image encoder and a 3D shape encoder--with a single scalar weight per scene, chosen without any ground-truth labels.
In open-vocabulary 3DIS, class scores are cosine similarities between a mask embedding and text anchor embeddings \cite{radford2021clip,zhou2023uni3d}.
Rather than relying on initial label predictions, our method leverages the mask proposals and 2D image embeddings from existing state-of-the-art pipelines \cite{takmaz2023openmask3d,nguyen2024open3dis,boudjoghra2024openyolo3d} and complements them with a frozen 3D shape encoder. 
We quantify each head's reliability by its sensitivity: how far it lifts the correct-class score above background noise, in units of its own background standard deviation.
To maximize the sensitivity of the fused prediction, we derive an optimal weight that depends solely on the first two moments of the score distributions.
The resulting rule coincides with the Bayes decision rule when the scores are Gaussian, and remains effective in practice when they deviate from it (Sec.~\ref{sec:experiments}).
Fig.~\ref{fig:architecture} illustrates the pipeline.
Sec.~\ref{subsec:SenseFuse} introduces the fusion rule. Sec.~\ref{subsec:labelfree} details the label-free weight estimation.

\subsection{Problem Formulation}
Let $\mathcal{P} = \{\mathbf{P}_n\}_{n=1}^{N} \in \mathbb{R}^{N \times 6}$ be an RGB-colored point cloud of a reconstructed indoor scene, where each $\mathbf{P}_n$ concatenates position and color.
Open-vocabulary 3DIS segments $Q$ binary instance masks
$\{\mathbf{M}^{q}\}_{q=1}^{Q}$, $\mathbf{M}^{q} \in \{0,1\}^{N}$,
and assigns each a class label $Y^{q} \in \mathcal{V}$ from a text vocabulary
$\mathcal{V} = \{c_1, \dots, c_K\}$, with $K$ set by the benchmark
\cite{rozenberszki2022language,straub2019replica,yeshwanth2023scannet++}.

We inherit the masks $\mathbf{M}^{q}$ and their 2D image embeddings $z^{q}_{2} \in \mathbb{R}^{d_2}$ from existing pipelines \cite{takmaz2023openmask3d,nguyen2024open3dis,boudjoghra2024openyolo3d}, and add a 3D shape embedding $z^{q}_{3} \in \mathbb{R}^{d_3}$ by passing the masked points $\mathcal{P}^{q} = \{\mathbf{P}_n \in \mathcal{P} \mid \mathbf{M}^{q}_{n} = 1\}$ through a frozen 3D shape encoder \cite{zhou2023uni3d,liu2023openshape}; the two are indexed by $h \in \{2,3\}$ for the 2D and 3D head.
Since $|\mathcal{P}^{q}|$ varies per mask while the encoder takes a fixed input budget, each subset is resampled to $N_0 = 10^4$ points, centered, and scaled by its farthest point into the unit sphere, following the encoder's pretraining convention \cite{zhou2023uni3d}.
Both heads score against the same class names, each embedded by its own text encoder into anchors $t_{h,c} \in \mathbb{R}^{d_h}$; all embeddings are $\ell_2$-normalized.
Our goal is to predict the class label $\hat{Y}^{q}$ for each mask by balancing the two heads according to their per-scene reliability (Sec.~\ref{subsec:SenseFuse}), without relying on ground-truth labels.

\subsection{SenseFuse}
\label{subsec:SenseFuse}

Our fusion builds on the observation that 2D and 3D heads exhibit disjoint failure patterns, rarely sharing identical wrong labels compared to 2D-only pairs (Table~\ref{tab:pairs}).
For a given mask $q$ and head $h \in \{2,3\}$, the raw cosine score for class $c \in \mathcal{V}$ is $s^{q}_{h,c} = \bigl\langle z^{q}_{h},\, t_{h,c} \bigr\rangle$.
Because the two encoders project features into distinct text-alignment spaces, these raw scores carry head-specific baselines and are not directly comparable.
We therefore center each mask's $K$ scores at their mean:
\begin{equation}
    S^{q}_{h,c} \;=\; s^{q}_{h,c} - \frac{1}{K}\sum_{c' \in \mathcal{V}} s^{q}_{h,c'} .
    \label{eq:scores}
\end{equation}

After centering, each head is characterized by the first two statistical moments: the mean elevation $\mu_h$ of the correct-class score above the background, and the standard deviation $\sigma_h$ of background (incorrect-class) scores.
We approximate both as shared across classes and masks within a scene.
This yields one scene-level reliability estimate per head, rather than a claim that category statistics are strictly identical.
The head's sensitivity is defined by the discriminability index of signal detection theory \cite{green1966signal},
\begin{equation}
    d'_h = \mu_h/\sigma_h \qquad (h \in \{2,3\}),
\label{eq:sensitivity}
\end{equation}
which measures how far head $h$ lifts the correct class above its own background noise.

\begin{figure}[t]
\centering
\includegraphics[width=0.8\columnwidth]{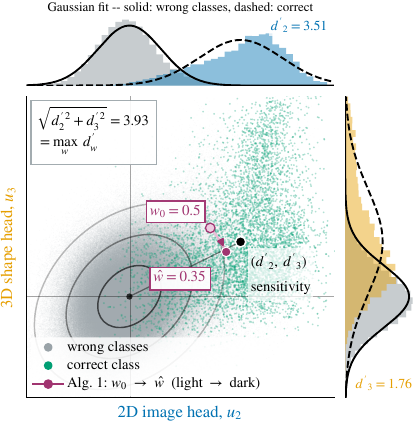}
\caption{\textbf{Visualization of weight estimation.}
     Scores $(u_2, u_3)$ ($u_h = S_h / \sigma_h$) for all matched mask--class pairs on Open-YOLO3D (EVA + Uni3D, ScanNet200), with marginal histograms and Gaussian fits; correct-class scores in color, wrong-class (background) scores in gray.
     The background ellipses show the measured noise correlation --- the tilt Eq.~\eqref{eq:law} deliberately ignores.
     The black dot marks the label-derived sensitivity $(d'_2, d'_3)$, reached from the background mean.
     The label-free iteration of Alg.~\ref{alg:fuse} (light to dark purple) converges to $\hat w = 0.35$, matching this direction.
}
\label{fig:mahalanobis}
\vspace{-3mm}
\end{figure}

We fuse the heads linearly with a single scalar $w$:
\begin{equation}
    S^{q}_{c}(w)
    = (1-w)S^{q}_{2,c} + wS^{q}_{3,c},
    \quad
    \hat{Y}^{q}
    = \arg\max_{c \in \mathcal{V}} S^{q}_{c}(w).
    \label{eq:fusion}
\end{equation}
Treating the two heads' background scores as uncorrelated (a design choice we justify below), the fused score exhibits an elevation of $(1-w)\mu_2 + w\mu_3$ and a background variance of $(1-w)^2\sigma_2^2 + w^2\sigma_3^2$, yielding a combined sensitivity of:
\begin{equation}
    d'_w = \frac{(1-w)\mu_2 + w\mu_3}{\sqrt{(1-w)^2\sigma_2^2 + w^2\sigma_3^2}} .
    \label{eq:dw}
\end{equation}
Maximizing this sensitivity by setting $\partial d_w'/ \partial w = 0$ provides the closed-form optimal weight
\begin{equation}
    w_{\mathrm{law}} = \frac{\mu_3 / \sigma_3^2}{\mu_2 / \sigma_2^2 + \mu_3 / \sigma_3^2}
    \label{eq:law}
\end{equation}
which indicates that the optimal weight for separating the correct class from the background depends only on these empirical per-head moments.
Defining $\pi_h = \mu_h/\sigma_h^{2}$, this weighting rule simplifies to $w_{\mathrm{law}} = \pi_3/(\pi_2+\pi_3)$, which naturally lies in $[0,1]$ whenever $\mu_h \ge 0$.

Under a Gaussian score model with a correct-class location shift $\mu_h$ and shared diagonal covariance, Eq.~\eqref{eq:law} coincides with Linear Discriminant Analysis (LDA) \cite{fisher1936use,hastie2009elements} and is Bayes-optimal under a uniform prior. 
Fig.~\ref{fig:mahalanobis} depicts these approximate Gaussian marginals.
Regardless, the deployed rule relies purely on the empirical moments and requires no strict distribution assumptions.
Although background scores correlate in practice ($\hat{\rho} \in [0.14, 0.26]$), explicitly correcting for this fails to improve accuracy and can produce inadmissible weights (Table~\ref{tab:rho}). 
We therefore deploy Eq.~\eqref{eq:law} as written, treating cross-modal correlation strictly as a diagnostic metric.

\subsection{Label-free Estimation}
\label{subsec:labelfree}
The optimal weight in Eq.~\eqref{eq:law} relies only on the moments $(\mu_h, \sigma_h)$. However, measuring the correct-class elevation $\mu_h$ requires knowing each mask's true label, which is exactly what is unknown in an open-vocabulary setting.
We therefore substitute soft responsibilities computed from the current fused scores for the missing labels, and iterate.
Hard $\arg\max$ pseudo-labels and hard confidence thresholds would make this iteration fragile, since discrete mask selections change discontinuously with $w$.
Following the soft responsibilities of Expectation--Maximization \cite{dempster1977em}, we soften both.
Alg.~\ref{alg:fuse} summarizes the complete procedure.

For a given weight $w$, let $\nu^{q} = \mathrm{std}_c\bigl(S^{q}_{c}(w)\bigr)$ be the spread of mask $q$'s fused scores, and $\tilde{m}^{q} = (m^{q} - \mathrm{med}(m)) / \mathrm{std}(m)$ its top-1 to top-2 margin $m^{q}$ standardized across the scene.
The soft class responsibility $P^{q}_c$, the background distribution $\Gamma^{q}_c$, and the mask subset weight $g^{q}$ are:
\begin{align}
    P^{q}_{c}
    &= \mathrm{softmax}_{c}\bigl(S^{q}_{c}(w)/(\tau_0\,\nu^{q})\bigr),
    \qquad
    \Gamma^{q}_{c}
    = \frac{1-P^{q}_{c}}{K-1}, \notag \\
    g^{q}
    &\propto \bigl[1+e^{-\beta\,\tilde m^{q}}\bigr]^{-1},
    \qquad
    \textstyle\sum_{q} g^{q} = 1 .
    \label{eq:soft}
\end{align}
Here, $P^{q}$ is a soft correct-class assignment, and $\Gamma^{q}$ relaxes the uniform background distribution by allocating the non-correct probability mass across the remaining $K-1$ classes.
The term $g^{q}$ provides a smooth relaxation for subset selection, where mask participation decays continuously as a function of the scene-median margin rather than through a step function.
The empirical moments are then responsibility-weighted averages across masks:
\begin{align}
    \hat\mu_h &= \sum_{q} g^{q}\Bigl(\sum_{c} P^{q}_{c}S^{q}_{h,c}
                 - \sum_{c} \Gamma^{q}_{c}S^{q}_{h,c}\Bigr), \notag \\
    \hat\sigma_h &= \sum_{q} g^{q}\Bigl[\sum_{c} \Gamma^{q}_{c}
                 \Bigl(S^{q}_{h,c}-\sum_{c'} \Gamma^{q}_{c'}S^{q}_{h,c'}\Bigr)^{2}\Bigr]^{1/2}.
    \label{eq:plugin}
\end{align}
Both hyperparameters are dimensionless, with $\tau_0$ scaling the row spread $\nu^{q}$ and $\beta$ scaling the standardized margin.
As a result, they are insensitive to the absolute score scale, allowing us to use the same default settings ($\tau_0 = 0.25$, $\beta = 8$) for every dataset and pipeline.
In the limit $\tau_0 \to 0$ and $\beta \to \infty$, $P^{q}$ converges to a one-hot $\arg\max$ vector and $g^{q}$ reduces to a hard median-margin indicator, confirming that Eq.~\eqref{eq:plugin} represents a continuous relaxation of the hard pseudo-label estimator.

Eq.~\eqref{eq:soft}--\eqref{eq:plugin} formulate a mapping where a scene-level weight $w$ induces soft assignments and subset weights, which in turn update the moments $(\hat{\mu}_h, \hat{\sigma}_h)$. 
Recalculating the fusion weight via Eq.~\eqref{eq:law} completes the iterative cycle.
Because the softened update is continuous in $w$ and maps $[0,1]$ into itself (lines 11--12 of Alg.~\ref{alg:fuse}), a fixed point exists by the Intermediate Value Theorem.
More importantly, within $R_{\max} = 12$ iterations, the process converges to the same dataset-level weight from every initialization $w_0 \in \{0, 0.25, 0.5, 0.75, 1\}$ in all tested settings, and to within $0.01$ on 302 of the 312 ScanNet200 scenes (Sec.~\ref{sec:experiments}).
Fig.~\ref{fig:mahalanobis} illustrates how the fusion weight converges to the fixed point ($\hat{w}=0.35$) from the default initialization $w_0 = 0.5$.
Initialization sensitivity at the per-scene level and the rationale for setting the default to $w_0 = 0.5$ are analyzed in Sec.~\ref{sec:experiments}.
Once converged, Alg.~\ref{alg:fuse} returns the final scene weight $\hat{w}$ and hard labels $\hat{Y}^{q}$. 
For mAP evaluation, the final instance confidence for each mask is assigned a fixed value of 1.0 following the convention of existing open-vocabulary pipelines~\cite{takmaz2023openmask3d,nguyen2024open3dis,boudjoghra2024openyolo3d} (Sec.~\ref{sec:experiments}).

\begin{algorithm}[t]
    \caption{Label-free sensitivity-weighted fusion}
    \label{alg:fuse}
    \begin{algorithmic}[1]
        \Require raw scores $s_2, s_3 \in \mathbb{R}^{Q\times K}$; init $w \gets w_0 = 0.5$; temperature $\tau_0 = 0.25$; sharpness $\beta = 8$; max iters $R_{\max} = 12$; tolerance $\varepsilon = 10^{-3}$; guard $\delta = 10^{-12}$
        \State Center rows: $S_h \gets s_h - \tfrac{1}{K}\sum_c s_{h,c}$ \Comment{Eq.~\eqref{eq:scores}}
        \For{$r = 1$ to $R_{\max}$}
        \State $F \gets (1{-}w)\,S_2 + w\,S_3$ \Comment{$F \equiv S(w)$, Eq.~\eqref{eq:fusion}}
        \State $\nu^{q} \gets \mathrm{std}_c(F^{q}_{c})$ \Comment{row spread}
        \State $P^{q}_{c} \gets \mathrm{softmax}_c\bigl(F^{q}_{c}/(\tau_0(\nu^{q}{+}\delta))\bigr)$
        \State $\Gamma^{q}_{c} \gets \tfrac{1-P^{q}_{c}}{K-1}$ \Comment{Eq.~\eqref{eq:soft}}
        \State $m^{q} \gets F^{q}_{(1)} - F^{q}_{(2)}$ \Comment{top-1/top-2 margin}
        \State $\tilde m^{q} \gets \bigl(m^{q}-\mathrm{med}(m)\bigr)/(\mathrm{std}(m)+\delta)$
        \State $g^{q} \propto \bigl[1+e^{-\beta \tilde m^{q}}\bigr]^{-1}$,\quad $\textstyle\sum_q g^{q} = 1$ \Comment{Eq.~\eqref{eq:soft}}
        \State $(\hat\mu_h,\hat\sigma_h) \gets \textsc{PlugIn}(S_h,P,\Gamma,g)$ \Comment{Eq.~\eqref{eq:plugin}}
        \State $\hat\pi_h \gets \max(\hat\mu_h/\hat\sigma_h^{2},\,0)$,\; $h \in \{2,3\}$
        \State $w' \gets \hat\pi_3/(\hat\pi_2+\hat\pi_3+\delta)$ \Comment{Eq.~\eqref{eq:law}}
        \State \textbf{if} $|w' - w| < \varepsilon$ \textbf{then} $w \gets w'$; \textbf{break} \Comment{converged}
        \State $w \gets w'$
        \EndFor
        \State $\hat w \gets w$ \Comment{final scene weight}
        \State $\hat Y^{q} \gets \arg\max_c\bigl[(1{-}\hat w)S_{2}+\hat w S_{3}\bigr]^{q}_{c}$ \Comment{hard labels}
        \State \Return $\hat w$, $\{\hat Y^{q}\}$
    \end{algorithmic}

\end{algorithm}

\section{Experiments}
\label{sec:experiments}

\begin{table*}[!t]
\caption{\textbf{SenseFuse accuracy}
    (\%) on three validation sets, each scored with its benchmark's official class list.
    $\hat w$: label-free dataset-level weight (Alg.~\ref{alg:fuse}), with sd of per-scene weights $\hat w_i$ in parentheses.
    $w^\star$: oracle weight --- the accuracy-maximizing per-dataset scalar, found on a 0.01 grid \emph{with} ground-truth labels.
    2D: standalone pipeline accuracy ($w{=}0$). @$\hat w$\,/\,@$\hat w_i$: accuracy under the dataset-level (bold, deployed) and per-scene weights --- the two are on par. rec: fraction of the oracle gain recovered by $\hat w$. $\dagger$: Open3DIS recomputed from raw cosines. OpenMask3D has no ScanNet++ proposals.
}
\label{tab:accuracy}
\centering
\footnotesize
\setlength{\tabcolsep}{2.5pt}
\begin{tabular}{@{}l rrrrr rrrrr rrrrr@{}}
\toprule
& \multicolumn{5}{c}{ScanNet200} & \multicolumn{5}{c}{Replica}
& \multicolumn{5}{c}{ScanNet++} \\
\cmidrule(lr){2-6}\cmidrule(lr){7-11}\cmidrule(lr){12-16}
Method & $\hat w\,(\mathrm{sd}\,\hat w_i)$ & $w^{\star}$ & 2D & @$\hat w$\,/\,@$\hat w_i$ & rec
       & $\hat w\,(\mathrm{sd}\,\hat w_i)$ & $w^{\star}$ & 2D & @$\hat w$\,/\,@$\hat w_i$ & rec
       & $\hat w\,(\mathrm{sd}\,\hat w_i)$ & $w^{\star}$ & 2D & @$\hat w$\,/\,@$\hat w_i$ & rec \\
\midrule
OY (EVA) & 0.352 (0.09) & 0.35 & 48.4 & \textbf{53.0}\,/\,52.9 & 99\% & 0.467 (0.08) & 0.31 & 61.2 & \textbf{71.2}\,/\,69.1 & 93\% & 0.472 (0.08) & 0.57 & 44.2 & \textbf{69.3}\,/\,70.1 & 91\% \\
OY (CLIP)  & 0.231 (0.10) & 0.24 & 41.1 & \textbf{48.0}\,/\,48.0 & 97\% & 0.352 (0.07) & 0.18 & 65.5 & \textbf{71.2}\,/\,72.7 & 67\% & 0.352 (0.13) & 0.44 & 50.2 & \textbf{70.5}\,/\,70.5 & 93\% \\
OpenMask3D   & 0.219 (0.09) & 0.30 & 29.4 & \textbf{33.9}\,/\,33.9 & 81\% & 0.376 (0.07) & 0.16 & 60.4 & \textbf{67.6}\,/\,69.1 & 67\% & --- & --- & --- & --- & --- \\
Open3DIS     & 0.236$^\dagger$ (0.09) & 0.21$^\dagger$ & 37.5$^\dagger$ & \textbf{44.6}$^\dagger$\,/\,43.5$^\dagger$ & 100\%$^\dagger$ & 0.163 (0.05) & 0.15 & 51.1 & \textbf{58.4}\,/\,58.2 & 97\% & 0.085 (0.04) & 0.12 & 68.0 & \textbf{71.8}\,/\,71.6 & 96\% \\
\bottomrule
\end{tabular}
\end{table*}
\begin{figure*}[t]
\centering
\includegraphics[width=.90\textwidth]{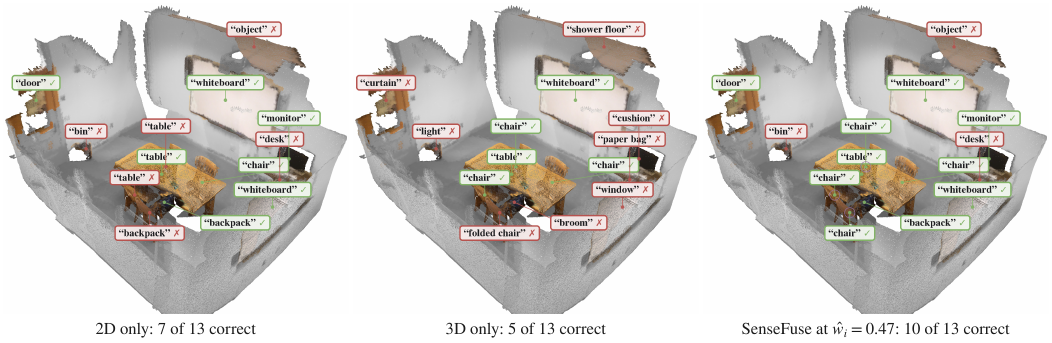}
\caption{\textbf{Qualitative results on a single scene.}
    Instance labeling on ScanNet200 scene0578\_00 by Open-YOLO3D (EVA, 2D only), Uni3D (3D only), and SenseFuse (green: correct, red: incorrect; labeled instances keep their color, while the rest of the room is shown in gray). 
    While the 2D and 3D heads fail on 6 and 8 instances respectively, SenseFuse ($\hat w_i=0.47$) correctly predicts 10 of 13 masks without degrading any existing correct labels.
}
\label{fig:room}
\vspace{-3mm}
\end{figure*}

\subsection{Setup}
We evaluate on three open-vocabulary 3D instance segmentation benchmarks: the official validation sets of ScanNet200~\cite{rozenberszki2022language} (312 scenes, 198 instance classes) and ScanNet++~\cite{yeshwanth2023scannet++} (50 scenes, 84 instance classes), alongside the standard 8-scene evaluation split of Replica~\cite{straub2019replica} (8 scenes, 48 instance classes) established by OpenMask3D~\cite{takmaz2023openmask3d}.
To isolate the impact of label assignment from spatial proposal quality, we adopt binary mask proposals and the per-mask 2D evidence (mask embeddings or image crops) of three state-of-the-art pipelines, Open-YOLO3D~\cite{boudjoghra2024openyolo3d}, OpenMask3D~\cite{takmaz2023openmask3d}, and Open3DIS~\cite{nguyen2024open3dis}. 
SenseFuse replaces only their final class assignments. 
Consequently, each pipeline's native labeling serves as the controlled baseline, ensuring that our evaluation of labeling accuracy is not confounded by spatial proposal recall.
The 2D head uses CLIP ViT-L/14~\cite{radford2021clip} (CLIP) mask embeddings for OpenMask3D and mask-pooled CLIP point features for Open3DIS.
For Open-YOLO3D, which scores image crops directly, we report results using both EVA02-CLIP-E/14+~\cite{sun2023eva} (EVA) and CLIP.
The 3D head utilizes Uni3D-giant~\cite{zhou2023uni3d} (Uni3D) as the default 3D shape encoder, with OpenShape PointBERT-ViT-g/14~\cite{liu2023openshape} (OS) evaluated in our modality-pairing study (Table~\ref{tab:pairs}).
All experiments are conducted on NVIDIA TITAN~V GPUs (12\,GB), with every model fitting on a single GPU.

We observe that standard Average Precision (AP) is heavily influenced by submission rules and confidence scoring strategies rather than pure classification accuracy. 
For example, with identical class scores, binary masks, and fixed confidence of $1.0$, evaluating only the top-1 prediction per proposal~\cite{nguyen2024open3dis,takmaz2023openmask3d} versus retaining the top-600 entries from the flattened proposal--class matrix~\cite{boudjoghra2024openyolo3d} yields $23.9$ versus $28.3$ mAP, respectively (Open-YOLO3D on ScanNet200, Table~\ref{tab:ap}).
Since SenseFuse modifies only proposal labels while leaving masks and confidences unchanged, its AP gains are also bounded by mask proposal recall.
Therefore, we adopt labeling accuracy as our primary metric: $\text{Acc} = \frac{1}{|\mathcal{Q}_{0.5}|} \sum_{q \in \mathcal{Q}_{0.5}} \mathbb{I}(\hat{Y}^{q} = Y^{q})$, matching each proposal to its highest-overlap ground-truth instance ($\text{IoU} \ge 0.5$; $6{,}534$ masks for Open-YOLO3D/OpenMask3D and $28{,}224$ masks for Open3DIS on ScanNet200).
We additionally report AP under both submission rules for comparability (Table~\ref{tab:ap}).

Because published baselines use inconsistent evaluation protocols, we re-evaluate all methods under this unified protocol.
Open3DIS releases only post-softmax, mask-filtered probabilities; thus, we re-run its official codebase to extract pre-softmax cosine scores on unfiltered proposals and evaluate them in the centered space of Eq.~\eqref{eq:scores} (marked with $\dagger$).

\begin{table}[!t]
\caption{\textbf{SenseFuse AP performance}
    under two standard submission rules: top-1 prediction per proposal (top-1) vs.\ top-600 entries of the flattened proposal--class matrix (flat); all proposals carry fixed confidence $1.0$. 
    Bold: improvement over 2D only.
    $\dagger$: Open3DIS from our re-run raw cosines.
}
\label{tab:ap}
\centering
\small
\setlength{\tabcolsep}{3.4pt}%
\begin{tabular}{ll rr rr rr}
\toprule
& & \multicolumn{2}{c}{ScanNet200} & \multicolumn{2}{c}{Replica}
& \multicolumn{2}{c}{ScanNet++} \\
\cmidrule(lr){3-4}\cmidrule(lr){5-6}\cmidrule(lr){7-8}
Method & Setting & top-1 & flat & top-1 & flat & top-1 & flat \\
\midrule
\multirow{2}{*}{OY (EVA)}
 & 2D only  & 23.9 & 28.3 & 17.5 & 18.3 & 6.1 & 6.9 \\
 & $+$Uni3D & \textbf{24.1} & 28.2 & \textbf{19.1} & \textbf{18.4} & \textbf{7.0} & \textbf{7.5} \\
\midrule
\multirow{2}{*}{OY (CLIP)}
 & 2D only  & 19.5 & 27.1 & 15.9 & 17.9 & 6.3 & 6.8 \\
 & $+$Uni3D & \textbf{20.3} & \textbf{27.5} & \textbf{16.2} & \textbf{18.4} & \textbf{7.4} & \textbf{7.3} \\
\midrule
\multirow{2}{*}{OM3D}
 & 2D only  & 15.5 & 24.4 & 15.8 & 18.0 & --- & --- \\
 & $+$Uni3D & \textbf{17.1} & \textbf{25.8} & \textbf{16.3} & \textbf{18.5} & --- & --- \\
\midrule
\multirow{2}{*}{O3DIS}
 & 2D only  & 15.3$^\dagger$ & 22.4$^\dagger$ & 11.6$^\dagger$ & 10.0$^\dagger$ & 11.2 & 10.5$^\dagger$ \\
 & $+$Uni3D & \textbf{16.6}$^\dagger$ & \textbf{23.6}$^\dagger$ & \textbf{14.3}$^\dagger$ & \textbf{14.5}$^\dagger$ & \textbf{12.0} & \textbf{12.4}$^\dagger$ \\
\bottomrule
\end{tabular}
\end{table}

\begin{figure}[t]
\centering
\includegraphics[width=\columnwidth]{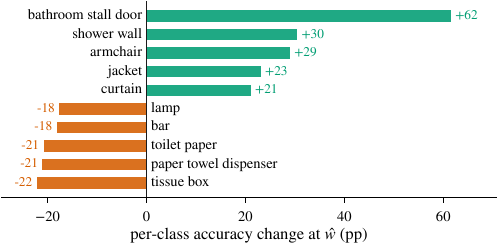}
\caption{\textbf{Top-5 most improved and degraded classes.}
    Per-class accuracy change (percentage points) at the deployed $\hat w$ for Open-YOLO3D (EVA + Uni3D) on ScanNet200, over classes with $\ge 10$ matched masks.
    Every degraded class has a median point count under $1{,}000$.
}
\label{fig:delta}
\vspace{-3mm}
\end{figure}

\subsection{Main Results}

Table~\ref{tab:accuracy} reports labeling accuracy across eleven (pipeline, dataset) settings. 
SenseFuse improves accuracy in every setting by a median of $7.1$ points (ranging from $+3.8$ to $+25.1$ points). 
The label-free weight recovers a median $93\%$ (minimum $67\%$) of the performance gain available at the oracle weight $w^{\star}$. 
This oracle is the single per-dataset weight that maximizes accuracy, selected on a $0.01$-spaced grid using ground-truth labels.

The dataset-level estimated weight $\hat{w}$ lands on an accuracy plateau close to that of the oracle, and it deviates from the mean of the per-scene weights $\hat{w}_i$ by at most $0.04$. 
We formally define oracle recovery as $\mathrm{rec} = \bigl(\mathrm{Acc}(\hat w)-\mathrm{Acc}_{0}\bigr)/\bigl(\mathrm{Acc}(w^{\star})-\mathrm{Acc}_{0}\bigr)$, where $\mathrm{Acc}_0$ is the 2D-only accuracy at $w{=}0$.
A bootstrap over scenes puts every ScanNet200 and ScanNet++ gain above zero, while Replica's eight scenes resolve only one of four.

To align with standard conventions, we also evaluate SenseFuse using the AP metric (Table~\ref{tab:ap}).
Although AP scores vary depending on submission rules even within identical pipelines, SenseFuse improves performance in 21 of 22 settings, achieving gains up to $+4.5$ mAP.

Fig.~\ref{fig:delta} shows class-wise accuracy changes.
Geometry-distinctive classes (e.g., bathroom stall doors, shower walls) gain up to $+62$ percentage points, while small objects defined primarily by texture or context (e.g., tissue boxes, paper towel dispensers) degrade.
This degradation is concentrated in sparse masks: on ScanNet200, only the lowest point-count quartile degrades ($-2.1$ pp in the 1st quartile, $+8.4$ pp in the 3rd).
Small masks ($<1{,}000$ points) constitute $28\%$ of ScanNet200 but only $0.7\%$ of ScanNet++, aligning with ScanNet++'s superior overall gain ($+25.1$ vs.\ $+4.5$ pp).

\begin{table}[t]
\caption{\textbf{Pair complementarity}
    on Open-YOLO3D (ScanNet200, 6{,}534 masks). 
    Acc$_1$, Acc$_2$: each head's standalone accuracy.
    $\rho$: mean per-mask Spearman rank correlation between the two heads' class score predictions.
    Ident.\: fraction of identical wrong labels.
    Resc.\: rescue rate (first head wrong, second right).
    Acc: fused accuracy. 
}
\label{tab:pairs}
\centering
\footnotesize
\setlength{\tabcolsep}{3.5pt}
\begin{tabular}{@{}ll rrrrrr@{}}
\toprule
Pair & Modality & Acc$_1$ & Acc$_2$ & $\rho$ & Ident. & Resc. & Acc \\
\midrule
EVA $+$ CLIP  & 2D$+$2D & 48.4 & 41.1 & 0.677 & 41.4\% & 7.8\% & $49.6$ \\
Uni3D $+$ OS  & 3D$+$3D & 16.1 &  3.6 & 0.394 &  4.4\% & 2.5\% & $12.4$ \\
\midrule
EVA $+$ Uni3D & 2D$+$3D & 48.4 & 16.1 & 0.226 &  5.9\% & 7.5\% & $\mathbf{53.0}$ \\
CLIP $+$ Uni3D& 2D$+$3D & 41.1 & 16.1 & 0.124 &  6.0\% & 9.0\% & $\mathbf{48.0}$ \\
EVA $+$ OS    & 2D$+$3D & 48.4 &  3.6 & 0.209 &  2.4\% & 1.6\% & $49.3$ \\
CLIP $+$ OS   & 2D$+$3D & 41.1 &  3.6 & 0.079 &  2.2\% & 2.2\% & $42.8$ \\
\bottomrule
\end{tabular}
\end{table}
\begin{table}[t]
\caption{\textbf{Ablation on cross-modal independence}
    of the correlation-corrected weight $w_\rho = \sigma_2(d'_3 - \rho d'_2) / \bigl[ \sigma_3(d'_2 - \rho d'_3) + \sigma_2(d'_3 - \rho d'_2) \bigr]$ against Eq.~\eqref{eq:law} on ScanNet200.
    $\hat\rho$: background score residual Pearson correlation estimated label-free via Alg.~\ref{alg:fuse} (distinct from the rank correlation $\rho$ in Table~\ref{tab:pairs}).
    Acc ($\Delta$): accuracy under $w_\rho$ (and change relative to $\hat w$). Inad.\: fraction of scenes yielding inadmissible weights ($w_\rho \notin [0, 1]$).
}
\label{tab:rho}
\centering
\small
\setlength{\tabcolsep}{5pt}
\begin{tabular}{@{}l rrr@{}}
\toprule
Setting & $\hat\rho$ & Acc ($\Delta$) & Inad. \\
\midrule
OY (EVA)  & 0.26 & $50.1$ ($-2.8$) & 20\% \\
OY (CLIP)   & 0.17 & $45.5$ ($-2.4$) & 9\% \\
OpenMask3D    & 0.18 & $32.1$ ($-1.8$) & 11\% \\
Open3DIS      & 0.14$^\dagger$ & $43.7^\dagger$ ($-0.9^\dagger$) & 3\%$^\dagger$ \\
\bottomrule
\end{tabular}
\vspace{-3mm}
\end{table}

\subsection{Why 2D--3D Fusion Works}
Table~\ref{tab:pairs} summarizes the performance of all four heads across every 2D/3D combination on ScanNet200.
Although CLIP is a stronger standalone 2D head than Uni3D (CLIP $41.1$, Uni3D $16.1$), its fusion with EVA yields only a $+1.2$ point gain, whereas Uni3D achieves a $+4.5$ point improvement.
Moreover, both heads exhibit a comparable rescue rate (CLIP $7.8\%$, Uni3D $7.5\%$).
Regarding error patterns, the two 2D image encoders share $41.4\%$ of identical incorrect labels among joint errors, whereas the 2D--3D pair shares only $5.9\%$.
This disparity aligns with the rank correlation of the two heads' class scores, with $\rho=0.677$ for the 2D pair compared to $\rho \in [0.079, 0.226]$ for the 2D--3D pairs.
Likewise, cross-modal pairing ensures the background noise correlation remains small enough ($\hat \rho \in [0.14, 0.26]$, Table~\ref{tab:rho}) to support the zero cross-covariance approximation in Eq.~\eqref{eq:law}.
As demonstrated in Table~\ref{tab:rho}, attempting to explicitly correct for the residual correlation fails to improve accuracy and can even yield inadmissible fusion weights.
However, each encoder's accuracy is also important.
For instance, OpenShape is the least correlated partner, but its low performance limits gains to $+0.9$ and $+1.6$.

Qualitatively, Fig.~\ref{fig:room} demonstrates how the two encoders complement each other to improve overall labeling accuracy.
The 2D head struggles with adjacency, misclassifying two chairs near the table as ``table'' and one next to a backpack as ``backpack''. 
The 3D head corrects the first two, while fusion successfully recovers the third (mislabeled in 3D as ``folded chair'').
Conversely, the 3D head misidentifies ``door'' as ``curtain'', ``chair'' as ``folded chair'', ``whiteboard'' as ``window'', ``backpack'' as ``broom'', and ``monitor'' as ``cushion''. 
These categories share coarse 3D geometries and require 2D texture or contextual cues for disambiguation, allowing fusion to correctly preserve the 2D predictions.
For the four masks that both encoders miss, their predicted wrong labels never coincide, perfectly aligning with our low shared-error rate. Finally, two of the three errors surviving fusion are near-synonyms (``bin'' for ``trash can'', ``desk'' for ``table'') that lie beyond either head's resolution.

\begin{figure}[t]
\centering
\includegraphics[width=\columnwidth]{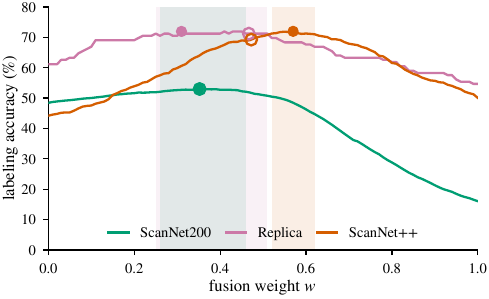}
\caption{\textbf{Labeling accuracy across fusion weights}
    for Open-YOLO3D (EVA + Uni3D) across three datasets. 
    Filled circles: oracle weight $w^\star$; open circles: deployed weight $\hat w$ (near-coincide on ScanNet200); shaded bands: region within 1\,pp of peak accuracy.
}
\label{fig:westimate}
\vspace{-3mm}
\end{figure}

\subsection{Weight Estimation and Ablation}
To evaluate the iteration's robustness to initialization, we sweep $w_0 \in \{0, 0.25, 0.5, 0.75, 1.0\}$.
With a naive hard $\arg\max$ pseudo-label update, the estimated weights differ by more than $0.01$ across initializations on $69$ of the $312$ ScanNet200 scenes (Open-YOLO3D, EVA).
The softmax relaxation in SenseFuse suppresses the variation to only $10$ scenes while keeping labeling accuracy within $0.3$ points.
Sweeping $\tau_0$ ($20\times$ range) and $\beta$ ($64\times$ range) yields median accuracy spans of only $0.7$ and $0.2$ points (worst $2.2$ and $0.7$), supporting a universal default across all settings.

Although the softened update guarantees the existence of a fixed point, Fig.~\ref{fig:westimate} illustrates that the exact location is not critical.
The estimated weight $\hat w$ consistently falls close to a broad accuracy plateau, where the region within $1$ point of the peak performance spans $0.06$--$0.27$ across configurations.
Even where $\hat w$ lands outside that region, as on ScanNet++, it still recovers 91\% of the oracle gain.
Nevertheless, this broad plateau is not shared across pipelines and datasets ($\hat w$ spans $0.085$--$0.472$). A single constant weight tuned jointly on all three validation sets recovers a mean $81\%$ of the oracle gain against $89\%$ for $\hat w$; in fact, a fixed $w{=}0.30$ drives ScanNet++/Open3DIS below its 2D-only accuracy.

Finally, Table~\ref{tab:rule} compares alternative fusion rules. The best score-level alternative (PoE) performs within $0.2$ points of our linear rule, while decision-level routing loses $1.2$ to $1.6$ points, proving that fusing scores before the \(\arg\max\) operation is essential. Furthermore, Open-YOLO3D's original sparse detection counts fall outside our continuous cosine model. This discrete space limits the 3D head to a minor Dirichlet prior (\(+1.0\)), which is why our main evaluation relies on its dense cosine variants.

\begin{table}[t]
\caption{\textbf{Ablation on the fusion rules}
    on Open-YOLO3D (EVA + Uni3D, ScanNet200). 
    Acc ($\Delta$): fused labeling accuracy and gain over the corresponding baseline row.
    Each cosine-space rule is evaluated label-free at its own estimated weight $\hat w$.
}
\label{tab:rule}
\centering
\small
\setlength{\tabcolsep}{4pt}
\begin{tabular}{@{}llr@{}}
\toprule
Rule & Formulation & Acc ($\Delta$) \\
\midrule
\multicolumn{3}{@{}l}{\textit{Score level (Centered cosine)}} \\
\textit{Baseline} & 2D only (EVA) & $48.4$ \\
Linear (\textbf{ours}) & weighted sum, Eq.~\eqref{eq:fusion} & $\mathbf{53.0}$ ($\mathbf{+4.5}$) \\
PoE     & log-linear pool of softmaxes & $52.8$ ($+4.3$) \\
MaxPool & $\max(s_2, s_3)$ per class & $48.4$ ($+0.0$) \\
Rank    & Borda count & $32.0$ ($-16.4$) \\
\multicolumn{3}{@{}l}{\textit{Decision level (after the $\arg\max$)}} \\
Router  & larger standardized margin wins & $46.9$ ($-1.6$) \\
Veto    & 3D vetoes 2D by rank & $47.2$ ($-1.2$) \\
\midrule
\multicolumn{3}{@{}l}{\textit{Vote space (Open-YOLO3D's own label scores)}} \\
\textit{Baseline} & Open-YOLO3D original labels & $48.6$ \\
Dirichlet & 3D shape as prior, $\alpha_0{=}1$ & $49.6$ ($+1.0$) \\
\bottomrule
\end{tabular}
\vspace{-3mm}
\end{table}

\subsection{Runtime Overhead}
We measured the per-scene runtime of the 3D shape encoder on Open-YOLO3D (EVA and Uni3D on ScanNet200), timing each stage individually on an NVIDIA TITAN~V GPU (12\,GB) averaged over the entire $312$-scene validation set.
The baseline pipeline spends $0.9$\,s on proposal generation and $24.6$\,s on 2D labeling ($25.4$\,s total).
SenseFuse adds $4.7$\,s for 3D encoding and $3.5$\,ms (maximum $11$\,ms) for weight estimation (Alg.~\ref{alg:fuse}).
Because 3D encoding depends solely on mask proposals, it can run concurrently with 2D labeling when a second GPU is available (EVA $9.4$\,GB, Uni3D $3.9$\,GB), expanding the critical path by only $3.5$\,ms.
Even under strict single-GPU serial execution, it adds merely $4.7$\,s ($18\%$), confirming the practical efficiency of SenseFuse.

\section{Conclusion}
\label{sec:conclusion}
In this work, we propose SenseFuse, a training-free framework that incorporates native 3D shape information into the labeling stage of open-vocabulary 3D instance segmentation without distillation. 
Motivated by the disjoint failure patterns of 2D images and 3D shape encoders, SenseFuse selects the scene-level fusion weight that maximizes a label-free sensitivity measure to balance the two heads. 
Across three pipelines and three benchmarks, SenseFuse consistently improves labeling accuracy. 
It recovers the vast majority of the oracle fusion gain at a per-scene estimation cost of just milliseconds.

SenseFuse currently fits one weight per scene and cannot account for instance-specific differences in modality reliability. This limitation naturally motivates future work on instance-level adaptation. 
Moreover, performance gains remain bounded by the weaker head, meaning that integrating stronger 3D shape encoders will directly raise the accuracy ceiling. 
Finally, extending this closed-form rule beyond two heads serves as a promising direction for future research.

\bibliographystyle{IEEEtran}
\bibliography{refs}

\end{document}